\documentclass[12pt]{article}
\usepackage{arxiv}
\usepackage[utf8]{inputenc}
\usepackage[T1]{fontenc}
\usepackage[greek,english]{babel}
\usepackage{amsmath}
\usepackage{hyperref}
\usepackage{url}
\usepackage{booktabs}
\usepackage{amsfonts}
\usepackage{graphicx}
\usepackage{subfigure}
\usepackage{natbib}
\usepackage{multirow,enumitem}
\usepackage{array}
\usepackage{algorithm}
\usepackage{authblk}
\usepackage[noend]{algpseudocode}

\newcolumntype{L}[1]{>{\raggedright\let\newline\\\arraybackslash\hspace{0pt}}m{#1}}
\newcolumntype{C}[1]{>{\centering\let\newline\\\arraybackslash\hspace{0pt}}m{#1}}
\newcolumntype{R}[1]{>{\raggedleft\let\newline\\\arraybackslash\hspace{0pt}}m{#1}}

\title{ORPHEAS: A Cross-Lingual Greek–English Embedding Model for Retrieval-Augmented Generation}

\author[1,2]{\href{https://orcid.org/0000-0002-3996-3301}{Ioannis E. Livieris}}
\author[1]{Athanasios Koursaris}
\author[1]{Alexandra Apostolopoulou}
\author[1]{Konstantinos Kanaris}
\author[1]{Dimitris Tsakalidis}
\author[1]{George Domalis}

\affil[1]{Novelcore, Athens, GR 10436\\
	\texttt{\{koursaris,domalis,apostolopoulou,kanaris,tsakalidis\}@novelcore.eu}}

\affil[2]{Department of Business Administration \& Organization Administration\\
	 University of Peloponnese, Kalamata,\\ 
	 GR 24100 \texttt{livieris@uop.gr}}

\renewcommand{\shorttitle}{\normalsize{ORPHEAS: A Cross-Lingual Greek–English Embedding Model for Retrieval-Augmented Generation}}

\hypersetup{
	pdftitle={ORPHEAS: A Cross-Lingual Greek–English Embedding Model for Retrieval-Augmented Generation},
	pdfsubject={cs.CL, cs.AI},
	pdfauthor={Ioannis E. Livieris, Athanasios Koursaris, Alexandra Apostolopoulou, Konstantinos Kanaris  Dimitris Tsakalidis, George Domalis},
	pdfkeywords={Embedding model, knowledge graph, Greek language, retrieval-augmented generation},
}

\begin{document}
    \maketitle

\begin{abstract}
Effective retrieval-augmented generation across bilingual Greek--English applications requires embedding models capable of capturing both domain-specific semantic relationships and cross-lingual semantic alignment. Existing multilingual embedding models distribute their representational capacity across numerous languages, limiting their optimization for Greek and failing to encode the morphological complexity and domain-specific terminological structures inherent in Greek text. In this work, we propose ORPHEAS, a specialized Greek--English embedding model for bilingual retrieval-augmented generation. ORPHEAS is trained with a high quality dataset generated by a knowledge graph-based fine-tuning methodology which is applied to a diverse multi-domain corpus, which enables language-agnostic semantic representations. The numerical experiments across monolingual and cross-lingual retrieval benchmarks reveal that ORPHEAS outperforms state-of-the-art multilingual embedding models, demonstrating that domain-specialized fine-tuning on morphologically complex languages does not compromise cross-lingual retrieval capability. \\ \\
	*** This paper has been accepted for presentation at \textit{Engineering Applications and Advances of Artificial Intelligence 2026 (EAAAI'26) }. Cite:  Livieris, I.E., Koursaris, A., Apostolopoulou, A., Kanaris, K.,
	Tsakalidis, D., Domalis, G. ORPHEAS: A Cross-Lingual Greek–English Embedding Model for Retrieval-Augmented Generation. In \emph{Engineering Applications and Advances of Artificial Intelligence}. ***
\end{abstract}


\section{Introduction}

Information retrieval (IR) in bilingual settings introduces compounded challenges, which extend beyond those encountered in monolingual systems. When one of the languages is morphologically complex, such as Greek \citep{koutsikakis2020greek}, inflectional variation produces multiple surface realizations of identical semantic concepts, undermining the practical applicability of conventional IR architectures based on lexical matching algorithms \citep{kadhim2019term}. This challenge is further amplified in cross-lingual retrieval scenarios, where queries and documents may reside in different languages, demanding that embedding models simultaneously encode morphological nuance and language-agnostic semantic alignment. General-purpose multilingual models, while demonstrating strong performance across high-resource languages, exhibit documented grammatical structure bias toward English-like syntactic patterns \citep{conneau2020emerging}, systematically disadvantaging morphologically complex, lower-resource languages such as Greek.

Retrieval-Augmented Generation (RAG) \citep{lewis2020retrieval} has emerged as a prominent architecture for knowledge-intensive Natural language processing (NLP) tasks, integrating Large Language Models (LLM) generative capabilities with retrieval mechanisms to mitigate hallucination and improve factual accuracy. Nevertheless, its effectiveness is fundamentally constrained by the quality of the underlying embedding model. This challenge is especially pronounced in Greek--English systems, where the model must simultaneously handle Greek morphological variation while maintaining robust semantic alignment across language boundaries. For the Greek academic and industrial community, which operates in a predominantly bilingual Greek--English environment, the availability of high-quality bilingual embeddings is of major importance. Despite this need, Greek remains an undeserved language in the NLP literature since existing multilingual models distribute their representational capacity across numerous languages \citep{wang2024multilingual,zhang2024mgtegeneralizedlongcontexttext}, limiting their optimization for Greek-specific morphological structures and domain-specific semantic relationships. Therefore, the development of specialized bilingual Greek--English embedding models constitutes a scientific and practical necessity for the Greek research community \citep{Themida}.

The development of high-quality embedding models for RAG typically involves a fine-tuning stage usually over domain-specific data, where the model learns to align query and document representations in a shared semantic space. Traditional fine-tuning methodologies for embedding models rely on synthetic query-document pair generation from document chunks \citep{bonifacio2022inpars,neeser2025quote}. While adequate for general domains, these approaches rely on surface-level textual patterns and simple positive-negative pair formulations, failing to capture the structured semantic relationships and hierarchical dependencies that characterize specialized corpora. Recently, \cite{Themida} attempted this limitation by treating a Knowledge Graph (KG) as a primary source for training data generation and proposed Themida, an advanced embedding model for Greek legal RAG systems. Rather than relying exclusively on synthetic query generation, their proposed approach extracts atomic facts, entity-relation contexts and grounded questions directly from knowledge graphs constructed over Greek legal corpora, demonstrating that structured knowledge integration substantially enhances embedding quality for morphologically complex legal text. However, Themida operates exclusively within the narrow scope of Greek legal documents, leaving the broader Greek academic and industrial community without a specialized solution for diverse domains and bilingual Greek--English retrieval.

In this research, we propose a specialized Greek--English embedding model, named ORPHEAS\footnote{The model's name, \textit{ORPHEAS}, stands for Optimized Representations for Parallel Hellenic-English Unified Semantics, named after the legendary ancient Greek musician and poet Orpheus, symbolizing the model's capacity to bridge languages and domains through semantic harmony.}, for bilingual RAG applications. Building upon the knowledge graph-based fine-tuning methodology of Themida, ORPHEAS extends its scope to a diverse multi-domain corpus, which enables the model to retrieve relevant information regardless of whether the query and document languages match. It is worth noting that the primary motivation for the development of ORPHEAS is the needs of the Greek academic and industrial community and it is designed to serve researchers, engineers and practitioners working with Greek and English text across diverse domains in a bilingual environment. The contributions of this work are summarized as follows:

\begin{itemize}
	\item We develop ORPHEAS, a specialized Greek--English embedding model for RAG applications, capable of capturing domain-specific semantic structures and morphological nuances across diverse Greek corpora while maintaining robust cross-lingual alignment with English.
	\item The development of ORPHEAS is based on a knowledge graph-based fine-tuning methodology, enforced with a cross-lingual augmentation strategy, enabling ORPHEAS to perform monolingual and cross-lingual retrieval within a unified model.
	\item The presented experimental and statistical analysis demonstrate that ORPHEAS outperforms state-of-the-art multilingual embedding models across monolingual and cross-lingual benchmarks.
\end{itemize}

The remainder of this paper is organized as follows. Section~\ref{Sec:2} reviews the related work on text embeddings. Section~\ref{Sec:4} presents the development of ORPHEAS, including the datasets, the fine-tuning methodology as well as the model architecture and training procedure. Section~\ref{Sec:5} presents the experimental analysis and discusses the comparative performance of \textsc{Orpheas} against state-of-the-art embedding models. Finally, Section~\ref{Sec:6} concludes the paper and outlines directions for future research.

\section{Related work}\label{Sec:2}

Text embedding models constitute of primary importance in many industrial and scientific applications, enabling advances in semantic search~\citep{reimers-2019-sentence-bert}, information retrieval~\citep{bajaj2018msmarcohumangenerated} and Retrieval-Augmented Generation 
\citep{lewis2021retrievalaugmentedgenerationknowledgeintensivenlp}. The recent development of transformer-based architectures and specialized fine-tuning strategies has further extended these capabilities to low-resource and morphologically complex languages \citep{Themida}. In the rest of this section, we provide a comprehensive review of the foundational models and specialized approaches proposed in the literature with Greek language support.

\cite{wang2020minilm}, proposed MiniLM, a compact multilingual embedding model supporting 50 languages including Greek, which transfers self-attention knowledge from larger teacher models through knowledge distillation. The model achieves 99\% performance retention with 50\% parameter reduction, making it particularly suitable for production environments where computational efficiency is critical.

cite{song2020mpnet} proposed MPNet, a multilingual embedding model supporting 50 most frequently utilized languages including Greek, addressing the position discrepancy problem \cite{siino2024all} through a novel two-stream self-attention mechanism with position compensation. Fine-tuned on one billion sentence pairs, MPNet produces 768-dimensional embeddings with strong performance across semantic textual similarity tasks, establishing it as a competitive benchmark for multilingual retrieval.

\cite{zhang2024mgtegeneralizedlongcontexttext} proposed mGTE, a multilingual embedding framework addressing long-context processing through Rotary Position Encoding supporting 8192-token contexts and hybrid dense-sparse retrieval with cross-encoder reranking. mGTE was trained on 2.8 billion multilingual text pairs across 75 languages and achieves state-of-the-art performance on several benchmarks. Nevertheless its reliance on Wikipedia and web-crawled data may introduce language-distribution biases.

\cite{wang2024multilingual} introduced the E5 family of instruction-based multilingual embeddings, employing a two-stage training pipeline combining weakly-supervised contrastive pre-training on one billion multilingual pairs with supervised fine-tuning using mined hard negatives and knowledge distillation. Despite strong cross-lingual performance across 93 languages, E5 distributes its representational capacity broadly, potentially limiting its optimization for specific language pairs.

\cite{Themida} proposed Themida, a specialized embedding model for the Greek legal domain, which leverages a knowledge graph to generate high-quality fine-tuning training data. Specifically, a LLM is used to extract atomic facts, entities and grounded questions from legal corpora and the generated knowledge is structured through a domain-specific ontology. Next, the KG is traversed to generate a set of semantically coherent anchor-positive pairs for contrastive learning, demonstrating exceptional performance on administrative and legislative Greek text.

Despite the advances of the presented models, their applicability to specialized Greek--English retrieval remains limited. General-purpose multilingual models distribute their representational capacity across numerous languages, limiting their optimization for Greek morphological structures and domain-specific semantics, while Themida is restricted to the narrow scope of Greek legal text. ORPHEAS levarages a KG-based fine-tuning methodology with a cross-lingual augmentation strategy across diverse domains, enabling robust monolingual and cross-lingual retrieval within a unified model.

\section{Development of ORPHEAS}\label{Sec:4}

This section presents the development of ORPHEAS, a Greek--English embedding model for bilingual RAG systems. Initially, We describe the datasets used for training, followed by the knowledge graph-based fine-tuning methodology and finally present the model architecture and training procedure.

\subsection{Datasets}\label{Sec:4.1}

The development of ORPHEAS leverages a diverse collection of corpora spanning multiple domains. Since Greek is the primary target language and high-quality Greek resources are substantially harder to obtain than English ones, the corpus is weighted toward Greek, encompassing encyclopedic knowledge, parliamentary discourse, legal documentation, administrative jurisprudence and social media content. English is represented through MS-MARCO, a large-scale IR dataset for which abundant resources are readily available.

\begin{itemize}
	\item \textit{Greek Wikipedia (Wiki)}\footnote{\url{https://el.wikipedia.org}}: Encyclopedic articles covering diverse topics including history, science, culture and geography. The corpus provides general knowledge expressed in modern Greek with neutral and informative tone.
	
	\item \textit{Hellenic Parliament Proceedings (Parliament)}\footnote{\url{https://www.hellenicparliament.gr/koinovouleftikes-epitropes/synedriaseis}}: Official transcripts from parliamentary sessions featuring deliberations, debates and expert testimonies. The language combines formal parliamentary discourse with conversational elements, covering legislative proposals and policy discussions across various national domains.
	
	\item \textit{Government Gazette Series A ($\Phi$EK-A)}\footnote{\url{https://www.et.gr}}: Official legal publications comprising laws enacted by the Hellenic Parliament, acts of legislative content, presidential decrees and regulatory instruments. The documents employ formal legal Greek following strict legislative drafting conventions and represent primary sources of Greek statutory law.
	
	\item \textit{Council of State Decisions (CoS)}\footnote{\url{https://www.adjustice.gr}}: Full-text judicial opinions from Greece's supreme administrative court, covering constitutional, tax, environmental and urban planning law. The corpus features formal legal Greek rich in administrative jurisprudence and includes structured metadata such as decision numbers, dates and case identifiers.
	
	\item \textit{Greek-Reddit}\footnote{\url{https://huggingface.co/datasets/IMISLab/GreekReddit}}: A topic classification dataset comprising 6534 posts with titles and topic labels collected from Greek subreddits \citep{mastrokostas2024social}. This dataset provides contemporary informal Greek language usage and social discourse patterns.
	
	\item \textit{MS-MARCO}\footnote{\url{https://microsoft.github.io/msmarco/}}: A large-scale English information retrieval dataset containing diverse queries and passages from web documents \citep{bajaj2016ms}. The passages are processed through the same knowledge graph construction pipeline as the Greek corpora, enabling extraction of entities, atomic facts and grounded questions, while the existing queries are directly used as anchors. Notice that only the training split is used for fine-tuning dataset generation.
\end{itemize}

\subsection{Knowledge Graph-Based Fine-Tuning Methodology}\label{Sec:4.2}

For developing ORPHEAS, we adopted the knowledge graph-based fine-tuning methodology proposed by \cite{Themida}. This approach leverages structured knowledge extraction to generate semantically enriched training data through contrastive learning. In detail, the methodology employs a domain-specific ontology $O$, which structures the knowledge graph around five core concepts: \textsc{Document}, \textsc{Chunk}, \textsc{Entity}, \textsc{Question} and \textsc{Atomic\_Fact}. The ontology defines hierarchical relationships where documents are decomposed into chunks via the \textsc{has\_chunk} relation and each chunk is associated with its semantic components through \textsc{has\_entity}, \textsc{has\_atomic\_fact} and \textsc{has\_question} relations. Entities are further connected through domain-specific \textit{relation} edges, capturing the semantic network within the corpus. This structured representation enables systematic extraction of semantically coherent anchor-positive pairs for contrastive learning.

Algorithm~\ref{Algorithm:1} outlines the three-stage methodology for constructing the knowledge graph, generating the fine-tuning dataset and applying cross-lingual augmentation. The algorithm operates on document corpus $\mathcal{C}$, chunk size $s$ and sampling parameters $m_a$, $m_q$, $m_e$ (controlling atomic facts, questions and entity-based sentences per chunk), ultimately producing fine-tuning dataset $D$ of semantically aligned anchor-positive pairs. In Stage~I, the algorithm constructs knowledge graph $\mathcal{G}$ by segmenting documents into chunks and using large language models to extract entities, relationships, atomic facts and grounded questions, which are then mapped to the ontology and integrated into the graph. In Stage~II, the algorithm generates training pairs by randomly sampling anchors (atomic facts, questions and entity-based sentences) from $\mathcal{G}$ for each chunk and pairing them with their corresponding chunks as positives. In Stage~III, cross-lingual augmentation is applied by translating each Greek anchor to English and vice versa, producing additional anchor-positive pairs where the anchor and chunk reside in different languages; thereby, enabling ORPHEAS to learn language-agnostic semantic representations and effectively handle monolingual Greek, monolingual English and cross-lingual retrieval within a unified model.

\begin{algorithm}[!ht]
	\caption{Fine-Tuning Dataset Generation for ORPHEAS}\label{Algorithm:1}
	\begin{algorithmic}[1]
		\Statex \hspace{-.5cm}\textbf{Require:} 
		\Statex $\mathcal{C}$: Document corpus
		\Statex $O$: Domain ontology
		\Statex $s$: Chunk size
		\Statex $m_a$: Number of atomic facts per chunk
		\Statex $m_q$: Number of questions per chunk
		\Statex $m_e$: Number of entity-based sentences per chunk
		\Statex \hspace{-.5cm}\textbf{Return:} 
		\Statex Fine-tuning dataset $D$
		\Statex
		\State $\mathcal{G} = \emptyset$ \Comment{Stage I}
		\ForAll{document $d \in \mathcal{C}$}
		\State Segment $d$ into chunks: $\{c_1, \ldots, c_k\}$ of size $s$
		\ForAll{chunk $c_i$} 
		\State Extract entities $\mathcal{E}_i$, relationships $\mathcal{R}_i$, atomic facts $\mathcal{A}_i$ using LLM
		\State Generate questions $\mathcal{Q}_i$ grounded in $c_i$ using LLM
		\State Map $\mathcal{E}_i$, $\mathcal{R}_i$, $\mathcal{A}_i$, $\mathcal{Q}_i$ to ontology $O$
		\State Integrate mapped elements into knowledge graph $\mathcal{G}$
		\EndFor
		\EndFor
		\Statex
		\State $D = \emptyset$ \Comment{Stage II}
		\ForAll{document $d \in \mathcal{C}$}
		\ForAll{chunk $c_i \in d$}
		\State Retrieve from $\mathcal{G}$:
		\State \hspace{0.4cm} (i) $A_i$: randomly sample $m_a$ atomic facts linked to $c_i$
		\State \hspace{0.4cm} (ii) $Q_i$: randomly sample $m_q$ questions answerable by $c_i$
		\State \hspace{0.4cm} (iii) $E_i$: randomly sample $m_e$ sentences generated from 5 entities in $c_i$
		\ForAll{anchor $x \in A_i \cup Q_i \cup E_i$}
		\State Add $(x, c_i)$ to $D$ \Comment{Anchor-positive pair}
		\EndFor
		\EndFor
		\EndFor
		\Statex
		\ForAll{pair $(x, c_i) \in D$ with $x$ in Greek} \Comment{Stage III}
		\State Translate $x$ to English: $x_{\text{EN}} = \text{Translate}(x)$
		\State Add $(x_{\text{EN}}, c_i)$ to $D$ \Comment{Cross-lingual anchor-positive pair}
		\EndFor
		\ForAll{pair $(x, c_i) \in D$ from MS-MARCO}
		\State Translate $x$ to Greek: $x_{\text{GR}} = \text{Translate}(x)$
		\State Add $(x_{\text{GR}}, c_i)$ to $D$ \Comment{Cross-lingual anchor-positive pair}
		\EndFor
		\State \Return $D$
	\end{algorithmic}
\end{algorithm}

\subsection{Model Architecture and Training}\label{Sec:4.3}

ORPHEAS is built upon Multilingual-E5-base\footnote{\url{https://huggingface.co/intfloat/multilingual-e5-base}} \citep{wang2024multilingual}, a state-of-the-art multilingual text embedding model with a transformer architecture of 278M parameters. Specifically, this model is fine-tuned using the high quality dataset generated through the knowledge graph-based methodology described in Algorithm \ref{Algorithm:1}. The fine-tuning process employed Multiple Negatives Ranking Loss \citep{henderson2017efficient}, a contrastive learning objective, which trains the model to produce embeddings where anchors are positioned closer to their corresponding positive chunks in the embedding space while being distant from negative samples drawn from other documents in the batch.

This specialized fine-tuning enables ORPHEAS to capture the linguistic nuances, semantic structures and domain-specific terminology of academic and technical texts, while maintaining robust monolingual and cross-lingual retrieval capabilities.

\section{Experimental Analysis}\label{Sec:5}

In this section, we conduct a comprehensive evaluation of the proposed ORPHEAS model for RAG application and compare its performance against that of the most efficient and widely used embedding models: Microsoft's Multingual E5 \citep{wang2024multilingual}, MPnet \citep{song2020mpnet}, MiniLM \citep{reimers-2019-sentence-bert} and mGTE \citep{zhang2024mgtegeneralizedlongcontexttext}. To further assess the quality of the generated fine-tuning dataset independently of the model architecture, we include GEM-XLM-RoBERTa \citep{apostolopoulou2025forging} in our experimental analysis, a model pre-trained on a large Greek--English corpus, which is fine-tuned following the same high-quality dataset as ORPHEAS.

To compare the retrieval performance of the evaluated models, we employ two metrics at two retrieval depths: Accuracy at $k$ (Acc@$k$), which measures the proportion of queries where at least one relevant document appears within the top-$k$ retrieved results; and Normalized Discounted Cumulative Gain at $k$ (NDCG@$k$), which considers both the relevance and ranking position of retrieved documents. Both metrics are reported at $k \in \{3, 10\}$ as in \citep{Themida} to capture performance characteristics at different retrieval depths: shallow retrieval (top-3) emphasizes precision in surfacing the most relevant content, while extended retrieval (top-10) assesses the model's ability to maintain relevance across a broader set of candidates as in \citep{Themida}. In addition, to mitigate the influence of specific datasets on the evaluation process, we examine the hypothesis whether the observed performance differences among models are statistically significant. For this purpose, we employ the non-parametric Friedman Aligned-Ranks (FAR) test \citep{hodges2011rank} to globally rank the models based on their performance and the Li post-hoc test \citep{li2008two} with a significance level of $\alpha = 5\%$ to identify statistically significant performance differences \citep{livieris2024evaluation,kiriakidou2024mutual}.

\subsection{Retrieval Evaluation on Public Monolingual and Cross-Lingual Benchmarks}

Next, we evaluate the retrieval performance of all models on four public datasets, covering monolingual Greek retrieval, cross-lingual Greek--English retrieval and established English benchmarks, spanning diverse domains and task complexities:

\begin{itemize}
	\item \textit{Greek Civics QA}\footnote{\url{https://huggingface.co/datasets/ilsp/greek_civics_qa}} is a question-answering dataset containing 407 questions related to Greek civics and social studies education. The dataset includes questions in Greek along with comprehensive answers sourced from Greek educational materials, primarily from the Social and Political Education curriculum for Greek middle school. 
	
	\item \textit{TruthfulQA Greek}\footnote{\url{https://huggingface.co/datasets/ilsp/truthful_qa_greek}} is composed by 817 adversarial question-answer pairs with parallel versions in English and Greek. Two evaluation variants are employed: \textit{TruthfulQA} with questions and answers in the same language (Greek or English) and \textit{TruthfulQA}* with cross-lingual pairs where Greek questions are answered in English and vice versa, enabling assessment of both monolingual and cross-lingual model capabilities.
	
	\item \textit{MS MARCO}\footnote{\url{https://huggingface.co/datasets/microsoft/ms_marco}} is a large-scale English question answering benchmark \citep{bajaj2016ms}, comprising queries of diverse types, including descriptive, numeric, entity-based, location-based, and person-based. For evaluation purposes, 1000 query-answer pairs were randomly sampled from the held-out test split, with no overlap with the fine-tuning training data.
\end{itemize}

Table~\ref{Table:Results} presents the comparative performance of all embedding models across the four evaluation datasets. Clearly, ORPHEAS demonstrates consistently superior retrieval performance, established as the top-performing model in both monolingual and cross-lingual retrieval scenarios.
On the \textit{Greek Civics QA} dataset, ORPHEAS achieves the highest Acc@3 and Acc@10 scores while maintaining competitive NDCG scores, with mGTE marginally outperforming ORPHEAS on NDCG by a negligible margin. On the \textit{TruthfulQA} and \textit{TruthfulQA*} datasets, ORPHEAS  considerably outperforms all baseline models reporting 4\%--10\% amd 6\%--11\% higher Acc@3 score, respectively. Similar findings can be conducted for NDCG metric where the proposed model exhibits the best scores. On the English \textit{MS MARCO} benchmark, ORPHEAS delivers competitive results, marginally below mGTE and Multilingual E5, confirming that domain specialization for Greek does not compromise English retrieval capability.


Regarding GEM-XLM-RoBERTa, which shares the same fine-tuning dataset as ORPHEAS but relies on a less advanced backbone, its competitive performance relative to general-purpose baselines validates the quality of the proposed dataset independently of the underlying architecture. Notably, GEM-XLM-RoBERTa outperforms MPnet and MiniLM across most metrics, further confirming that the knowledge graph-based fine-tuning methodology contributes meaningfully to retrieval quality. Furthermore, the performance gap between GEM-XLM-RoBERTa and ORPHEAS, especially on cross-lingual retrieval, can be attributed to architectural limitations rather than the fine-tuning methodology itself.

\begin{table}[!ht]
	\centering
	\setlength{\tabcolsep}{2pt}
	\renewcommand{\arraystretch}{1}
	\caption{Retrieval performance comparison of embedding models across public monolingual and cross-lingual benchmarks}
	\begin{tabular}{l|l|cc|cc}
		\toprule
		\textbf{Dataset} & \textbf{Model} & \textbf{Acc@3} & \textbf{Acc@10} & \textbf{NDCG@3} & \textbf{NDCG@10} \\
		\midrule
		\multirow{5}{*}{\begin{tabular}[c]{@{}l@{}}Greek Civics\\QA\end{tabular}} 
		& Multilingual E5 & 92.62 & 97.54 & 87.25 & 89.25 \\
		& MPnet & 85.51 & 95.33 & 78.47 & 82.05 \\
		& MiniLM & 81.57 & 92.13 & 73.32 & 77.33 \\
		& mGTE & 93.36 & 98.03 & 88.08 & 89.80 \\
		& GEM-XLM-RoBERTa & 85.74 & 97.05 & 76.83 & 80.93 \\
		& ORPHEAS & 94.34 & 99.75 & 86.62 & 88.70 \\
		\midrule
		\multirow{5}{*}{TruthfulQA} 
		& Multilingual E5 & 82.12 & 84.94 & 80.49 & 81.54 \\
		& MPnet & 79.31 & 82.61 & 76.75 & 77.92 \\
		& MiniLM & 78.33 & 82.86 & 75.95 & 77.63 \\
		& mGTE & 81.39 & 84.57 & 79.86 & 81.05 \\
		& GEM-XLM-RoBERTa & 80.78 & 84.33 & 78.90 & 80.20 \\
		& ORPHEAS & 86.16 & 90.20 & 83.36 & 84.83 \\
		\midrule
		\multirow{6}{*}{TruthfulQA*} 
		& Multilingual E5    & 78.82 & 82.80 & 75.62 & 77.09 \\
		& MPnet         & 78.21 & 83.23 & 75.11 & 76.55 \\
		& MiniLM        & 78.09 & 82.68 & 74.92 & 76.58 \\
		& mGTE               & 78.21 & 83.90 & 74.71 & 76.36 \\
		& GEM-XLM-RoBERTa    & 77.89 & 83.31 & 74.13 & 75.49 \\
		& ORPHEAS            & 84.76 & 90.21 & 81.46 & 83.43 \\
		\midrule
		\multirow{5}{*}{MS MARCO} 
		& Multilingual E5 & 98.20 & 99.49 & 95.96 & 96.46 \\
		& MPnet & 73.57 & 83.95 & 67.39 & 71.22 \\
		& MiniLM & 70.33 & 80.52 & 64.34 & 68.11 \\
		& mGTE & 99.13 & 99.76 & 97.37 & 97.61 \\
		& GEM-XLM-RoBERTa & 96.39 & 98.98 & 93.15 & 94.13 \\
		& ORPHEAS & 97.74 & 99.33 & 95.42 & 96.01 \\
		\bottomrule
	\end{tabular}
	\label{Table:Results}
\end{table}

Table~\ref{Table:Statistical analysis} presents the statistical analysis based on FAR and Post-hoc Li tests for Acc@3 and NDCG@3 metrics, as shallow retrieval depth is considered the most stringent indicator of model precision. ORPHEAS exhibits the lowest FAR score across both metrics, confirming its superior retrieval performance. In addition, the Li post-hoc test reveals that for Acc@3, ORPHEAS significantly outperforms all competing models, while for NDCG@3 the performance differences between ORPHEAS, Multilingual E5 and mGTE are not statistically significant, suggesting comparable ranking quality among the top-performing models.

\begin{table}[!ht]
	\centering
	\setlength{\tabcolsep}{5pt}
	\renewcommand{\arraystretch}{1}
	\caption{Statistical analysis based on FAR and Post-hoc Li tests for (a) Acc@3 and (b) NDCG@3 metrics}
	\begin{tabular}{l|ccc}
		\toprule
		Model    & FAR & \multicolumn{2}{c}{\underline{Li post-hoc test}}\\
		& & APV & $H_0$\\
		\midrule  
		ORPHEAS          &  5.25 & ---    & ---                \\
		Multilingual E5  &  8.25 & 0.002 & Rejected           \\
		mGTE             &  8.63 & 0.001 & Rejected           \\
		GEM-XLM-RoBERTa  & 14.0 & 0.0 & Rejected           \\
		MPnet       & 18.88 & 0.0 & Rejected           \\
		MiniLM      & 20.0 & 0.0 & Rejected           \\
		\bottomrule
		\multicolumn{4}{c}{(a) Acc@3} \\
		\multicolumn{4}{c}{} \\
		\toprule
		Model    & FAR & \multicolumn{2}{c}{\underline{Li post-hoc test}}\\
		& & APV & $H_0$\\
		\midrule        
		ORPHEAS          &  6.50 & ---    & ---                    \\
		Multilingual E5  &  7.75 & 0.211 & Failed to reject        \\
		mGTE             &  8.25 & 0.099 & Failed to reject       \\
		GEM-XLM-RoBERTa  & 14.50 & 0.0 & Rejected               \\
		MPnet       & 18.25 & 0.0 & Rejected               \\
		MiniLM      & 19.75 & 0.0 & Rejected               \\
		\bottomrule
		\multicolumn{4}{c}{(b) NDCG@3}
	\end{tabular}\label{Table:Statistical analysis}
\end{table}

\subsection{Retrieval Evaluation on the Real-World DLBT Deliberation Platform}

To assess the practical applicability of ORPHEAS in a real-world setting, we conducted an additional evaluation using data derived from the DELIBERATE platform\footnote{\url{www.deliberate.gr}}. This platform has the attractive features to semantically group user comments into thematic clusters, each associated with a title and a descriptive summary \citep{Livieris2026}. The retrieval corpus comprises user comments, cluster titles and cluster summaries, providing a realistic and heterogeneous document collection. We constructed a dataset of 600 queries (300 monolingual and 300 cross-lingual) through an expert-driven annotation procedure involving two domain specialists in legal scholarship and computational linguistics, who independently constructed queries following structured guidelines designed to simulate authentic user retrieval behavior. This yields two evaluation settings: DLBT (monolingual), where queries and cluster descriptions share the same language and DLBT (cross-lingual), where queries and responses reside in opposite languages.

Table~\ref{tab:DLBT} presents the retrieval performance of all evaluated 
models on the DLBT datasets. In the monolingual setting, ORPHEAS 
consistently outperforms all competing models, achieving a 3\%--13\% 
improvement in Acc@3 over the baseline models. MPNet and MiniLM exhibit 
the weakest performance, reflecting their limited capacity for 
domain-specific Greek retrieval. As regards the cross-lingual setting, some interesting findings are revealed. Multilingual E5 undergoes a substantial 
performance degradation of approximately 15\% in Acc@3, despite its strong 
monolingual performance, whereas ORPHEAS maintains its leading position 
with a 3\%--13\% Acc@3 improvement over all competing models, further 
validating its effectiveness across linguistic settings. Notably, 
GEM-XLM-RoBERTa delivers competitive results in both settings, 
surpassing stronger baselines such as mGTE and Multilingual E5 in the 
cross-lingual configuration, which further corroborates the 
quality of the proposed fine-tuning dataset in inducing robust semantic 
representations regardless of the underlying architecture.

\begin{table}[!ht]
	\centering
	\setlength{\tabcolsep}{2pt}
	\renewcommand{\arraystretch}{1}
	\caption{Retrieval performance comparison of embedding models on the real-world DLBT deliberation platform}
	\begin{tabular}{l|l|cc|cc}
		\toprule
		\textbf{Dataset} & \textbf{Model} & \textbf{Acc@3} & \textbf{Acc@10} & \textbf{NDCG@3} & \textbf{NDCG@10} \\
		\midrule
		\multirow{6}{*}{\begin{tabular}[c]{@{}l@{}}DLBT\\(monolingual)\end{tabular}} 
		& Multilingual E5 & 90.53 & 94.74 & 88.79 & 90.32 \\
		& MPnet & 80.00 & 85.96 & 76.75 & 78.84 \\
		& MiniLM & 81.40 & 86.67 & 78.01 & 79.83 \\
		& mGTE & 88.77 & 92.28 & 84.74 & 86.02 \\
		& GEM-XLM-RoBERTa & 85.61 & 93.68 & 81.67 & 84.56 \\
		& ORPHEAS & 93.33 & 94.39 & 89.81 & 90.19 \\
		\midrule
		\multirow{6}{*}{\begin{tabular}[c]{@{}l@{}}DLBT\\(cross-lingual)\end{tabular}} 
		& Multilingual E5 & 75.60 & 82.00 & 72.83 & 75.18 \\
		& MPnet & 81.60 & 87.20 & 78.38 & 80.48 \\
		& MiniLM & 80.80 & 86.40 & 78.18 & 80.15 \\
		& mGTE & 83.60 & 92.00 & 81.43 & 84.53 \\
		& GEM-XLM-RoBERTa & 85.20 & 89.20 & 79.70 & 81.18 \\
		& ORPHEAS & 88.40 & 94.40 & 85.48 & 87.54 \\
		\bottomrule
	\end{tabular}
	\label{tab:DLBT}
\end{table}

\section{Conclusions and future research}\label{Sec:6}

In this work, we presented ORPHEAS, a specialized Greek--English model for bilingual RAG applications, developed through a knowledge graph-based fine-tuning methodology augmented with a cross-lingual strategy. The development of ORPHEAS is motivated by the absence of high-quality bilingual embedding solutions for the Greek academic and industrial community, which operates in a predominantly bilingual Greek--English environment where specialized NLP resources remain scarce. The proposed approach addresses this critical gap by enabling robust monolingual and cross-lingual retrieval within a unified model, without sacrificing English retrieval capability.

The experimental evaluation across public benchmarks and a real-world deliberation platform demonstrated that ORPHEAS outperforms state-of-the-art multilingual embedding models across all datasets and evaluation metrics, with statistically significant improvements in accuracy-based retrieval. The inclusion of GEM-XLM-RoBERTa as a secondary baseline further substantiates the quality of the proposed fine-tuning dataset, confirming that the knowledge graph-based methodology contributes meaningfully to retrieval quality, independent of the underlying architecture.

Our future work is concentrated on extending ORPHEAS to additional languages relevant to the Greek academic and industrial community, as well as 
exploring the effect of varying chunk sizes on retrieval performance;
thereby, broadening its applicability across diverse retrieval settings. 
Furthermore, we intend to evaluate the model on additional real-world 
datasets spanning a wider range of domains, which would further assess 
the generalization capacity and practical utility of ORPHEAS in 
operational deployment scenarios.

\subsubsection*{Acknowledgements} AWS resources were provided by the National Infrastructures for Research and Technology (GRNET) and funded by the EU Recovery and Resiliency Facility. This work received funding from the Horizon Europe research and innovation programme under Grant Agreement No. 101214389, project AIXPERT (An agentic, multi-layer, GenAI-powered backbone to make an AI system).

\bibliographystyle{unsrtnat}
\bibliography{bibliography}

\end{document}